\documentclass[journal]{IEEEtran}
\usepackage{cite}
\ifCLASSINFOpdf
\else
\fi
\usepackage{amsmath}
\usepackage{amssymb}
\usepackage[pdftex]{graphicx}
\usepackage{algorithmic}
\usepackage{tabularx}
\usepackage{booktabs}
\usepackage{multirow}
\usepackage{url}
\usepackage{color}
\begin{document}
\title{SDL-MVS: View Space and Depth\\ Deformable Learning Paradigm for Multi-View\\ Stereo Reconstruction in Remote Sensing}

\author{Yong-Qiang Mao, 
        Hanbo Bi,
        Liangyu Xu,
        Kaiqiang Chen,
        Zhirui Wang, 
        Xian Sun,
        Kun Fu
        % <-this % stops a space
\thanks{
This work was supported by the National Natural Science Foundation
of China under Grant 62331027 and Grant 62076241, and supported by the Strategic Priority Research Program of the Chinese Academy of Sciences, Grant No. XDA0360300. (Corresponding author: Xian Sun.) 

Yong-Qiang Mao, Hanbo Bi, Liangyu Xu, Xian Sun, and Kun Fu are with the Aerospace Information Research Institute, Chinese Academy of Sciences, Beijing 100190, China, the Key Laboratory of Network Information System Technology (NIST), Aerospace Information Research Institute, Chinese Academy of Sciences, Beijing 100190, China, the University of Chinese Academy of Sciences and the School of Electronic, Electrical and Communication Engineering, University of Chinese Academy of Sciences, Beijing 100190, China (e-mail: maoyongqiang19@mails.ucas.ac.cn; bihanbo21@mails.ucas.ac.cn; xuliangyu21@mails.ucas.ac.cn; sunxian@aircas.ac.cn; kunfuiecas@gmail.com).

Kaiqiang Chen and Zhirui Wang are with the Aerospace Information Research Institute, Chinese Academy of Sciences, Beijing 100190, China, the Key Laboratory of Network Information System Technology (NIST), Aerospace Information Research Institute, Chinese Academy of Sciences, Beijing 100190, China (e-mail: chenkaiqiang14@mails.ucas.ac.cn; zhirui1990@126.com).

}
}

\markboth{Manuscript to IEEE TGRS}%
{Shell \MakeLowercase{\textit{et al.}}: Bare Demo of IEEEtran.cls for IEEE Journals}

\maketitle



\begin{abstract}
Research on multi-view stereo based on remote sensing images has promoted the development of large-scale urban 3D reconstruction. However, remote sensing multi-view image data suffers from the problems of occlusion and uneven brightness between views during acquisition, which leads to the problem of blurred details in depth estimation. To solve the above problem, we re-examine the deformable learning method in the Multi-View Stereo task and propose a novel paradigm based on view Space and Depth deformable Learning (SDL-MVS), aiming to learn deformable interactions of features in different view spaces and deformably model the depth ranges and intervals to enable high accurate depth estimation. Specifically, to solve the problem of view noise caused by occlusion and uneven brightness, we propose a Progressive Space deformable Sampling (PSS) mechanism, which performs deformable learning of sampling points in the 3D frustum space and the 2D image space in a progressive manner to embed source features to the reference feature adaptively. To further optimize the depth, we introduce Depth Hypothesis deformable Discretization (DHD), which achieves precise positioning of the depth prior by adaptively adjusting the depth range hypothesis and performing deformable discretization of the depth interval hypothesis. Finally, our SDL-MVS achieves explicit modeling of occlusion and uneven brightness faced in multi-view stereo through the deformable learning paradigm of view space and depth, achieving accurate multi-view depth estimation. Extensive experiments on LuoJia-MVS and WHU datasets show that our SDL-MVS reaches state-of-the-art performance. It is worth noting that our SDL-MVS achieves an MAE error of 0.086, an accuracy of 98.9\% for $<0.6$m and 98.9\% for $<$3-interval on the LuoJia-MVS dataset under the premise of three views as input.

\end{abstract}


\IEEEpeerreviewmaketitle

\section{Introduction}
\IEEEPARstart{I}{n} an environment where remote sensing satellites are developing rapidly, remote sensing intelligent interpretation technology is widely used in fields such as semantic segmentation\cite{qian2023semantic}, target recognition\cite{yang2021rethinking, cheng2022anchor,li2023ogmn}, and three-dimensional reconstruction\cite{kunwar2020large,mao2023elevation}.
Among them, an increasing number of high-resolution optical remote sensing satellites with stereoscopic detection capabilities have promoted the development of 3D reconstruction\cite{de2014automatic,mao2022beyond,yu2021automatic,mao2023elevation} of the earth's surface and brought a solid data foundation, which has attracted widespread attention from researchers\cite{mao2022bidirectional,ozcanli2014automatic,sun2024gable}.
In recent years, depth estimation\cite{yao2018mvsnet,chen2020mvsnet++,gu2020cascade,liu2020novel} of Multi-View Stereo (MVS) images has been a popular research area.
However, 3D reconstruction of large geographical spatial ranges in remote sensing is mainly based on commercial software\cite{SURE-Aerial,Pix4D,ContextCapture}, and the commonly used method based on traditional dense matching consumes a lot of manpower and material resources.
In addition, large-scale 3D reconstruction of the earth's surface plays an important role in urban environmental monitoring\cite{yuan2020deep,li2020review}, disaster assessment\cite{hamdi2019forest,calantropio2021deep,bi2023not}, urban planning\cite{pan2020deep,huang2018urban}, autonomous driving\cite{grigorescu2020survey}, etc.
Therefore, the research on the multi-view stereo task in large-scale remote sensing scenes is urgent.

Most of the existing multi-view stereo methods focus on object-level reconstruction, such as MVSNet\cite{yao2018mvsnet}, Cas-MVSNet\cite{gu2020cascade}, UCS-Net\cite{cheng2020deep}, MVSFormer\cite{cao2022mvsformer } and so on.
MVSNet\cite{yao2018mvsnet} first proposed an end-to-end depth estimation network for multi-view stereo reconstruction, by constructing a depth hypothesis and using a differentiable homography matrix to generate cost volumes, and finally using multi-scale 3D U-Net to achieve subsequent feature decoding.
Considering that the memory and time costs of MVSNet will show a cubic growth trend when the input resolution increases, Cas-MVSNet\cite{gu2020cascade} proposed a multi-stage depth estimation scheme to construct the depth hypothesis range from coarse to fine gradually. However, object-level multi-view stereo methods are difficult to directly generalize to large-scale remote sensing scenes. This is because, unlike object-level multi-view image data, multi-view remote sensing images are easily affected by uneven illumination. What's more, there are many instances in the image, resulting in more target occlusion problems between views.
Therefore, in the field of remote sensing multi-view stereo research, Liu et al.\cite{liu2020novel} proposed the first aerial remote sensing dataset for large-scale MVS matching and earth surface reconstruction, and proposed the RED-Net network, using a recurrent encoder-decoder architecture to encode and decode cost volumes obtained from multi-view images. However, this method takes up a large amount of memory and has a slow training speed. For this reason, Li et al.\cite{li2023hierarchical} proposed a hierarchical deformable cascade MVS network, HDC-MVSNet, which simultaneously performs high-level resolution multi-scale feature extraction and hierarchical cost volume modules constructed to generate full-resolution depth with rich contextual information.
Although these methods have achieved excellent results, they still ignore the problem of detail blur caused by occlusion and uneven brightness between targets in remote sensing scenarios, which makes it difficult for existing methods to meet the needs of practical applications.
 
\begin{figure}[!t]
    \begin{center}
    \includegraphics[width=1.0\linewidth]{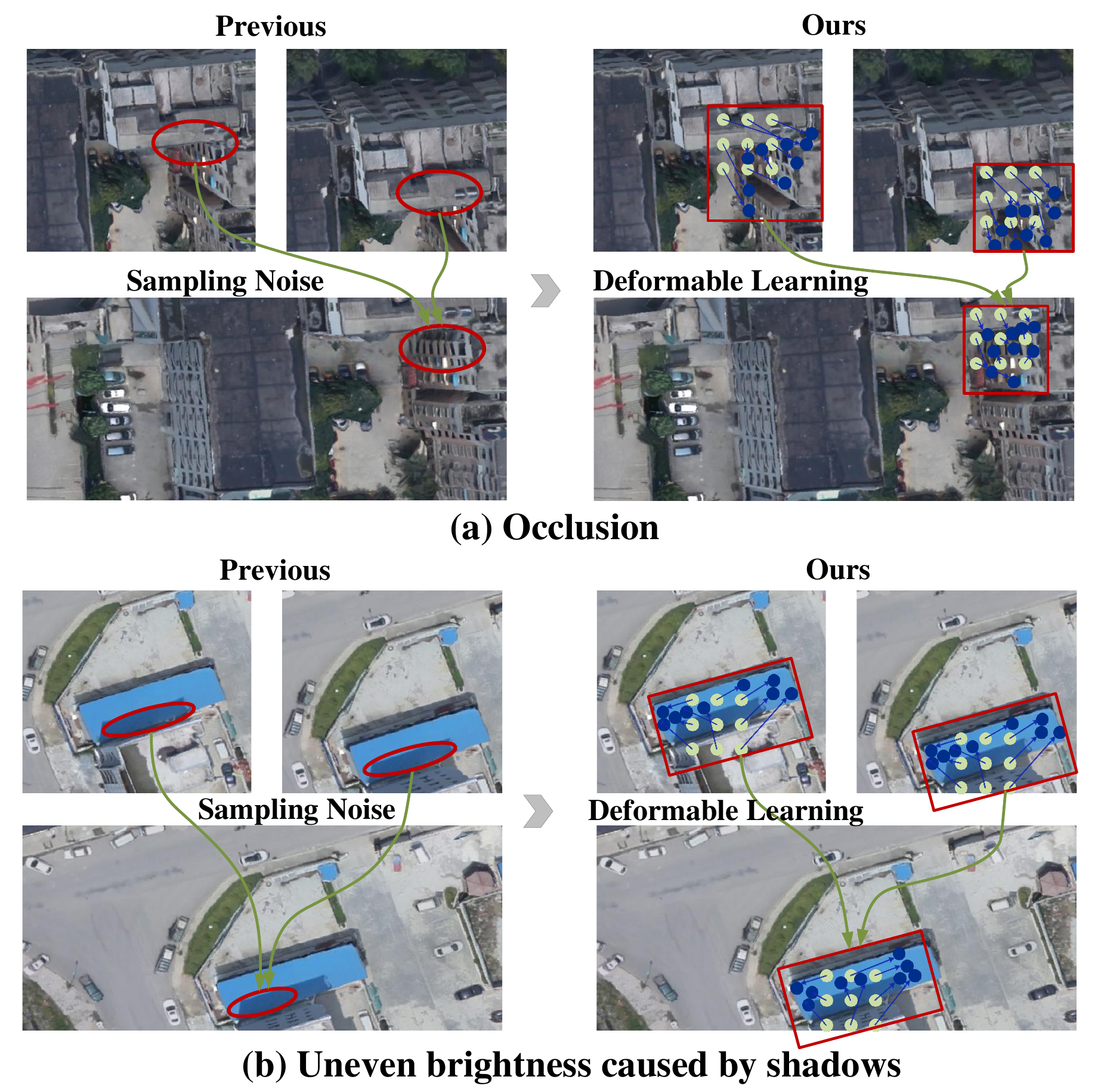}
    \end{center}
    \caption{The phenomenon of occlusion and uneven brightness. \textbf{(a) Occlusion.} Occlusion brings feature loss to the multi-view sampling process. Our method can adaptively learn similar features of nearby neighbors to reduce the impact of loss. \textbf{(b) Uneven brightness.} Uneven illumination caused by shadows or other factors causes different parts of the same structure of the same object to have different characteristics, which is not conducive to depth recovery. Our method can utilize adaptive supplementation of surrounding pixel features to address the negative impact of uneven illumination.}
    \label{1-motivation}
\end{figure}

From the remote sensing perspective, multi-view images often suffer from problems such as uneven brightness or occlusion between targets. Existing methods are limited in their generalization capabilities for the 3D reconstruction of large scenes, resulting in blurred details in the results of multi-view stereo tasks. Specifically, unlike the object-level three-dimensional reconstruction task in vision, remote sensing multi-view stereo collects data from high altitudes, so it will be affected by internal sensors or the external environment during the collection process. 
In large-scale scenes, objects with smooth surfaces such as buildings and water on the ground are easily affected by light reflection or shadows, resulting in uneven brightness of pixels corresponding to imaging. This phenomenon reduces the detailed information of the corresponding area to a certain extent, greatly affects the quality of the overall depth estimation, and leads to the problem of blurred details. 
In addition, in remote sensing scenarios, due to the limited tilt angle of the sensor, the side facades of buildings may be blocked by trees, billboards, and other objects, which results in insufficient information from multiple views. Only some views can capture the information of the building's side facade, making the corresponding area blurry. The problem of blurred details caused by uneven brightness and occlusion between views limits the accuracy improvement of multi-view stereo estimation.
As shown in Fig.~\ref{1-motivation}(a) and (b), the reason is that whether it is uneven brightness or occlusion between views, homography transformations essentially inevitably introduce noise into the generated cost volume features. 
Specifically, due to the occlusion phenomenon, when feature sampling is performed from multi-view images using camera pose constraints between views, the matching features of the target of interest should be sampled, but irrelevant features of the objects that impose occlusion are inevitably sampled. 
The inconsistency caused by feature sampling brings inevitable feature noise to the three-dimensional feature volume generated by homography transformation. However, the further forward propagation of the noise in 3D feature volumes transfers the error to the 3D cost volume, resulting in the accumulation of errors and ultimately affecting the depth estimation accuracy.
Similarly, uneven brightness will cause the feature at the corresponding position to deviate greatly in brightness from the unique texture of the object itself, which also leads to the introduction of noise in the three-dimensional feature volume generated during the homography transformation process, thereby affecting the generation of the cost volume.
Therefore, the problems of occlusion between targets and uneven brightness will lead to inconsistency in multi-view feature sampling, which will greatly affect the accuracy of depth estimation.
\begin{figure}[!t]
    \begin{center}
    \includegraphics[width=1.0\linewidth]{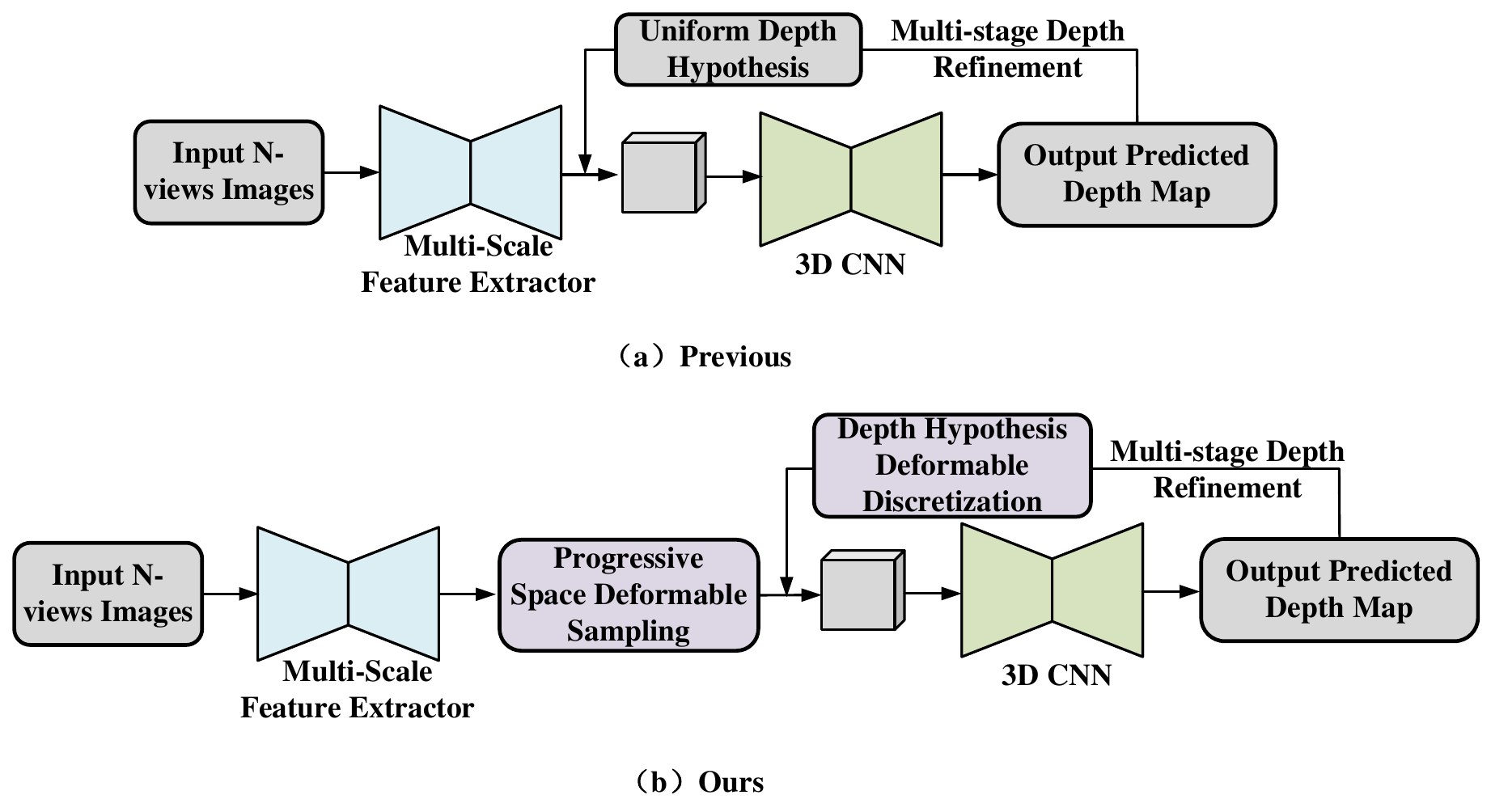}
    \end{center}
    \caption{Comparison of SDL-MVS with existing methods. Compared with (a) previous methods, (b) our method introduces a progressive space deformable sampling strategy embedded between the feature extractor and the cost volume generation process and a deformable discretization strategy of depth hypothesis, to solve the problems of occlusion and uneven brightness.}
    \label{2-comparemethod}
\end{figure}

To solve the above problems in the multi-view stereo task, we propose a Multi-View Stereo paradigm rooted in view Space and Depth deformable Learning (SDL-MVS) to achieve high-precision multi-view depth estimation.
To solve the problem of blurred details caused by occlusion and uneven brightness, as shown in Fig.~\ref{2-comparemethod}, a Progressive Space deformable Sampling (PSS) mechanism is proposed, which learns the sampling point offsets to sample view features in both three-dimensional frustum space and two-dimensional image space. 
The camera poses constraints between view images are adopted to perform regional adaptive aggregation of feature consistency to achieve sufficient feature interaction between the target area of the reference image and the corresponding area of the source image, reducing the adverse effects caused by irrelevant noise. 
% and solving the problem of information loss caused by occlusion and uneven brightness.
To further optimize the predicted depth, a Depth Hypothesis deformable Discretization (DHD) strategy is proposed which establishes the depth range and the depth interval deformable hypothesis. 
The range hypothesis is compressed according to the uncertainty of the depth range and the interval hypothesis is divided through a linear increasing discretization method according to the central depth, which eliminates outliers affected by photometry and constructs a more accurate depth prior range in multi-stage depth estimation.

Overall, the main contributions of our work can be summarized as:
\begin{itemize}
    \item We re-examine the deformable learning approach in the Multi-View Stereo task and propose a novel view Space and Depth deformable Learning paradigm (SDL-MVS) to solve the problems of occlusion and uneven brightness in multi-view remote sensing images to achieve high-precision depth estimation.
    \item We propose a Progressive Space deformable Sampling (PSS) mechanism, which achieves the aggregation of features from different views by progressively performing deformable learning of sampling points in the 3D frustum space and the 2D image space, effectively reducing the impact of feature noise caused by occlusion and uneven brightness.
    \item We introduce a Depth Hypothesis deformable Discretization (DHD) mechanism, which further optimizes multi-stage depth estimation and reduces the impact of undesirable noise by mining the uncertainty relationship of the depth range hypothesis and the discretization strategy of the depth interval hypothesis.
    \item Extensive experiments on popular benchmarks, LuoJia-MVS and WHU datasets, show that our SDL-MVS achieves new state-of-the-art performance whether taking 3-views or 5-views as input.
\end{itemize}

\section{Related Work}\label{sec:Related Work}
In recent years, works of the multi-view stereo task\cite{yao2018mvsnet,zhang2023geomvsnet,zhang2023vis,shi2023raymvsnet++,vats2024gc,chen2023costformer,khot2019learning,dai2019mvs2,mallick2020learning,liu2020novel,huang2018deepmvs,luo2020attention} are booming and applied to many fields. 
In this section, we introduce the methods based on fine feature extraction\cite{hartmann2017learned,im2019dpsnet,chang2018pyramid}, the methods based on cost volume generation\cite{guo2019group,im2019dpsnet,yi2020pyramid,xu2020learning}, and the methods based on depth hypothesis\cite{gu2020cascade,yang2020cost,yang2021cost2,cheng2020deep}.

\subsection{Methods Based on Fine Feature Extraction}
In multi-view stereo networks, fine deep feature extraction\cite{hartmann2017learned,im2019dpsnet,chang2018pyramid,xue2019mvscrf,chen2020mvsnet++,ramachandran2019stand,yan2020dense,yang2020fade,zhang2021long} is crucial.
In the early years, many works had begun to explore the use of deep neural networks to extract features from multi-view images.
To extract fine semantic features from multi-view images, Hartmann et al.\cite{hartmann2017learned} innovatively proposed a matching function embedded in multiple conjoined convolutional neural network branches and then generated similarity scores by mapping matching image patches. 
However, point-by-point depth estimation of this method requires multiple forward propagations according to the number of depth hypotheses, which results in a high computational cost and low inference speed.
2D convolutional neural networks (CNNs) have been shown to learn from many more efficient methods in vision tasks for multi-scale feature learning.
Im et al.\cite{im2019dpsnet} conducted multi-scale learning of multi-view features by introducing a typical spatial pyramid pooling structure to collect contextual information in different perceptual areas\cite{chang2018pyramid}. 
At the same time, the feature pyramid structure\cite{lin2017feature} in visual tasks is also embedded into the feature extractor\cite{xue2019mvscrf,gu2020cascade} of the multi-view network to learn multi-scale semantic features.
Furthermore, to improve the appearance robustness of object-level multi-view images, Chen et al.\cite{chen2020mvsnet++} embedded the instance normalization operation into the feature pyramid structure mentioned previously.
In addition, to ensure the feature resolution of the input, D$^2$HC-RMVSNet\cite{yan2020dense} designed a lightweight dense reception and expansion strategy aimed at learning multi-scale contextual information.
 
Existing feature extractors already have strong feature extraction capabilities. However, the generation of cost volumes in the multi-view stereo field is the core step of the network and a key factor in determining model performance. 
In this paper, we focus on how to use camera constraints between multi-view images to generate efficient 3D feature volumes.

\subsection{Methods Based on Cost Volume Generation}
Among the multi-view stereo methods based on deep learning, the most typical one is MVSNet\cite{yao2018mvsnet}, which earlier introduced the generation of cost volume into multi-view depth estimation.
Specifically, MVSNet employs the camera constraints between the multi-view images to perform homography transformation to generate the cost volume and uses 3D convolution to perform feature decoding on the generated cost volume.
In the multi-view stereo network, the generation of cost volume\cite{guo2019group,im2019dpsnet,yi2020pyramid,xu2020learning,yu2021attention,xu2020pvsnet,luo2019p,zhang2020visibility} is the core step of the entire depth estimation network. 
Most existing methods adopt variance calculation proposed by MVSNet to generate 3D cost volumes corresponding to multi-view images.
To extract a more accurate 3D cost volume, DPSNet\cite{im2019dpsnet} employs a plane sweep algorithm and averages all image pairs to construct a fine cost volume from depth features.
However, this cost volume generated using variance easily contains unnecessary redundant information of a certain view and has a large memory occupation cost, which limits further applications. 
Therefore, to construct a lightweight 3D cost volume, Xu et al.\cite{xu2020learning} construct an average group-wise correlation similarity measure between the reference and source image features. 
The correlation operation can perform correlation measurement in the form of explicit encoding, thereby making the generated 3D cost volume greatly reduce the burden of aggregation. 
To provide more effective semantic features for cost volume generation, Chen et al.\cite{chen2020mvsnet++} proposed MVSNet++ and introduced the attention mechanism into deep feature extraction and then integrated it into the generation process of cost volume. MVSNet++ improves the robustness of the generated 3D cost volume by adding pairs of cost volumes and introducing depth mask priors.

However, the features transmitted to the cost volume are extremely susceptible to noise in the multi-view stereo field of remote sensing. This is reflected in the fact that occlusion and uneven brightness will lead to greater noise in features from the specific view. 
Therefore, we focus on the deformable feature interaction between multiple views for reducing the noise of the feature volume before the cost volume to solve the problems of occlusion and uneven brightness.

\begin{figure*}[ht]
    \begin{center}
    \includegraphics[width=1.0\linewidth]{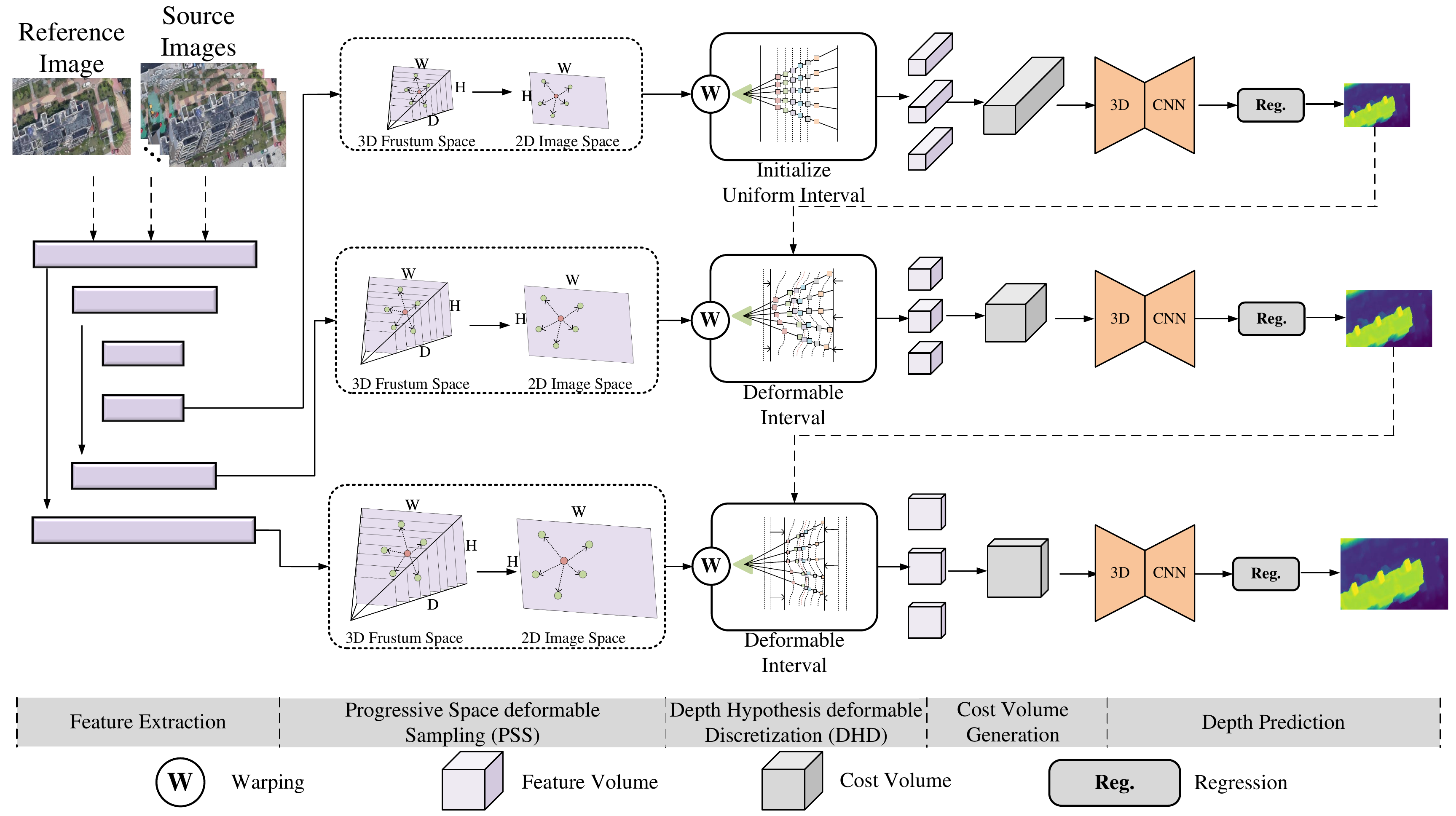}
    \end{center}
    \caption{The framework diagram of our view Space joint Depth deformable Learning paradigm (SDL-MVS). Our SDL-MVS consists of five processes, namely feature extraction, Progressive Space deformable Sampling (PSS), Depth Hypothesis deformable Discretization (DHD), cost volume generation, and depth prediction. The input data of the framework is the reference remote sensing image and source remote sensing images, and the output is the depth estimation map of the reference remote sensing image.}
    \label{3-framework}
\end{figure*}
\subsection{Methods Based on Depth Hypothesis}
Multi-stage depth estimation network\cite{gu2020cascade,yang2020cost,yang2021cost2,cheng2020deep} has been proven to be an effective structure for coarse-to-fine depth estimation due to its simplicity and efficiency.
CasMVSNet\cite{gu2020cascade} was proposed earlier, which performed coarse-to-fine optimization of the prediction range by setting a fixed number of depth hypotheses and corresponding depth intervals at each stage.
However, the fixed depth range hypotheses are insufficient to cover the ground-truth depth values and therefore cannot give a correct estimate.
Yi et al.\cite{yi2021ddr} introduced a distance estimation module to estimate depth uncertainty from a probability volume via a lightweight 2D CNN. For a reasonable estimation of distances, the training phase was re-examined using new depth ranges and input probability volumes.
Yang et al.\cite{yang2020cost,yang2021cost2} designed a specific criterion to set the depth range hypothesis of CVPMVSNet, which was determined by the corresponding pixel offset in the source view. Unlike other coarse-to-fine methods, CVPMVSNet is performed on an image pyramid instead of a feature pyramid, and the network parameters are shared on the pyramid level.

However, although these methods take into account the uncertainty of the depth hypothesis, they still suffer from the fact that they only focus on the adaptive learning of the depth range hypothesis and ignore the impact of uneven settings of the depth interval on depth estimation. In this paper, we focus on the adaptive learning of both depth range and interval and propose a deformable learning mechanism to construct fine depth hypothesis priors.

\section{Proposed Method}\label{sec:Proposed Method}
\subsection{Overview}\label{sec:Method Overview}
We revisit deformable feature learning in the multi-view stereo field and introduce in detail the proposed view space and depth deformable learning paradigm. 
Fig.~\ref{3-framework} gives the flowchart of the entire framework, which mainly includes the following parts: multi-scale feature extractor, Progressive Space deformable Sampling (PSS), Depth Hypothesis deformable Discretization (DHD), cost volume generation, and depth prediction.

Given multi-view remote sensing images $\{I_i\}_{i=1}^{N}$ ($N$ is the number of views), the network first performs deep feature learning of the images through the multi-scale feature extractor, which can construct deep features of multiple views at different scales for multi-scale cost volume learning. 
After obtaining multi-scale deep feature maps which can be found in Fig.~\ref{3-framework}, these features at the same scale and their corresponding camera internal and external parameter matrices are simultaneously fed into the progressive space deformable sampling module, aiming to aggregate consistent features between multiple views by learning adaptive offsets of sampling points in different spaces of source images.
After the aggregation of features between views, the features at each stage are used to generate a cost volume, which is based on the pre-assumed depth range and interval. What's more, the internal and external camera parameter matrix corresponding to each view also plays a big role.
In each stage, our SDL-MVS uses 3D convolutions to encode and decode features of the cost volume for subsequent depth estimation and probability distribution prediction.

\subsection{Progressive Space Deformable Sampling}
There is a certain overlap of pixels between the input multi-view remote sensing images, and every two images are closely related by the camera's intrinsic and extrinsic parameter matrices. In complex real-world scenes, objects will inevitably be occluded in different views, which will lead to blurring in the predicted depth map. 
To address this issue, we propose a Progressive Space deformable Sampling (PSS) mechanism to achieve feature complementation between multiple views.

Given the features $\{F_{i,1}\}_{i=1}^{N}$, $\{F_{i,2}\}_{i=1} ^{N}$ and $\{F_{i,3}\}_{i=1}^{N}$ obtained by the multi-scale feature extractor, our SDL-MVS performs deformable feature aggregation on the corresponding multi-view feature maps at each stage. 
Taking the $k$th stage as an example, $F_{1,k}$ and $\{F_{i,k}\}_{i=2}^N$ represent the reference and source features of the $k$th stage after the feature extractor, respectively.
Due to the spatial inconsistency between different views, we first construct the grid matrix $G$ under the reference view and then implement the grid matrix through the matrix transformation relationship between different source views and the reference view.
\begin{figure*}[ht]
    \begin{center}
    \includegraphics[width=0.8\linewidth]{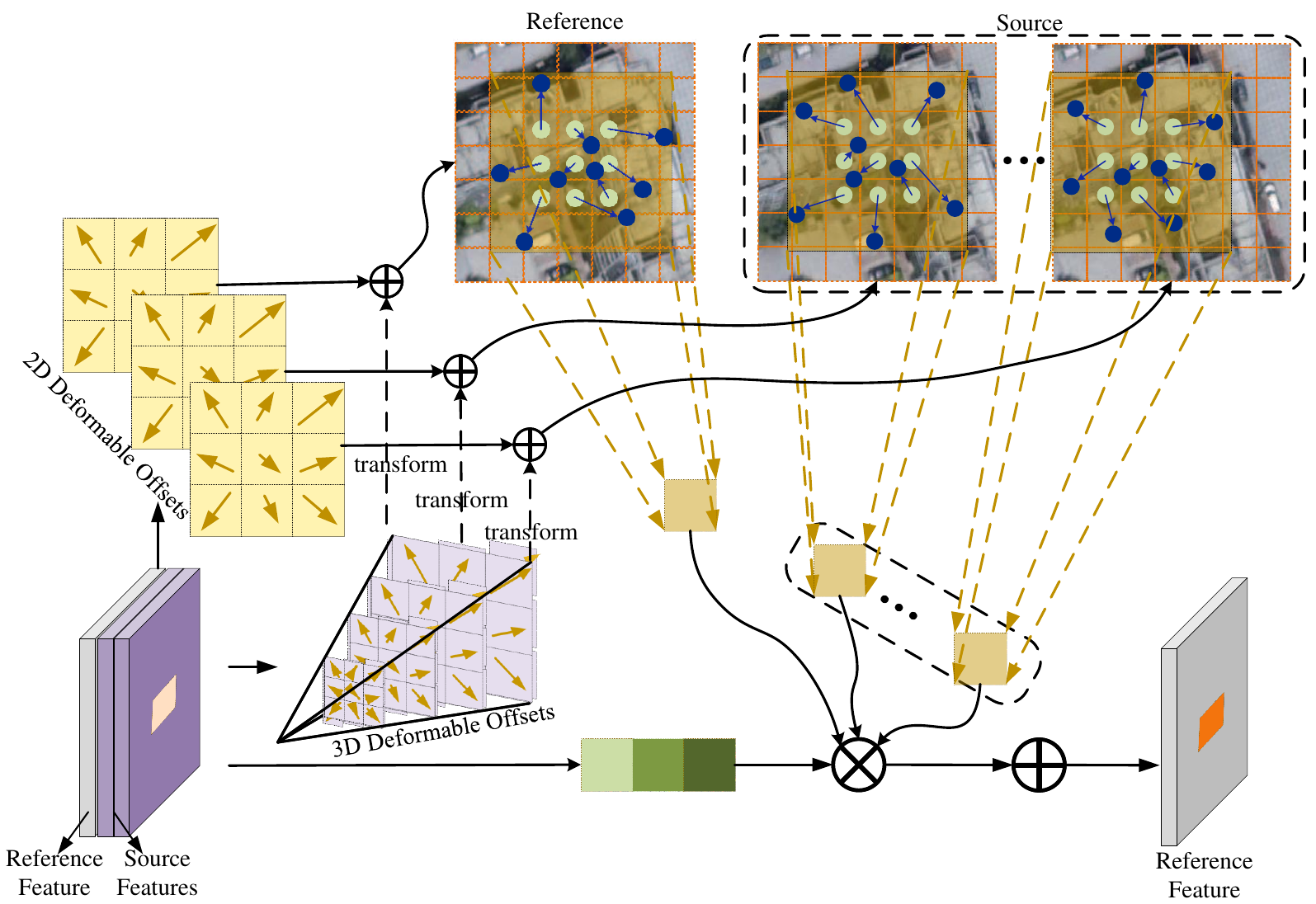}
    \end{center}
    \caption{The schematic of the proposed Progressive Space deformable Sampling (PSS). The input of the module is the reference features and source features obtained after the multi-scale feature extractor. The reference features are updated through progressive space feature sampling, and new reference features are output after feature aggregation.}
    \label{4-PSS}
\end{figure*}
 
In complex actual scenes, drones or aircraft equipped with multi-view cameras have occlusion, uneven brightness, and other unavoidable problems when collecting data. 
This leads to possible large differences in objects near pixels in different views with the same features.
We realize the positioning of pixels with the same features from different views through the consistency of features. 
Furthermore, to better learn the features around the projected pixel points, we set up an offset point learning strategy, which aims to achieve sampling learning by learning the offset vector around the central projection point.

Based on the constructed grid matrix, we first perform offset learning of sampling points in the three-dimensional frustum space. Here, the grid matrix $G$ is a three-dimensional grid matrix composed of height $H^{3D}$, width $W^{3D}$, and depth $Z^{3D}$. 
The coordinates of each point are composed of $(x,y,z)$.
To be able to perform adaptive offset learning in the three directions of length, width, and depth in the space of each view, we use the reference and source features to learn the three-dimensional offset, which is expressed as:
\begin{equation}
    \Delta G^{3d}_i = conv(cat(F_1, F_i))
\end{equation}
Among them, $conv$ and $cat$ represent convolution and concatenate operations, respectively, and $\Delta G^{3d}_i$ is the grid offset in the three-dimensional space learned by the network. Based on this, the grid data of the $i$th view after offset is $G+\Delta G^{3d}_i$.

Given the intrinsic and extrinsic camera parameters matrices $\{K_i, T_i\}$ of the $i$th view, the transformation matrix of $4\times 4$ dimensions from the current reference view to the specified $i$th source view is given by:
\begin{equation}
    H_i = K_iT_iT_1^{-1}K_1^{-1}
\end{equation}

Next, transform the grid matrix of the reference view into the specified $i$th source view to obtain grid matrices $G_i^{2D}$ of the $i$th source view after projection, which can be expressed as:
\begin{equation}
    G_i^{2d} = H_i\cdot (G+\Delta G^{3d}_i)
\end{equation}

As shown in Fig.~\ref{4-PSS}, after learning the offset in the three-dimensional frustum space, the grid data needs to be further converted into coordinates in the two-dimensional image space, and the coordinate of offsets in the two-dimensional space is also learned in the multi-view space, which can be expressed as:
\begin{equation}
    \Delta G^{2d}_i = conv(cat(F_1, F_i))
\end{equation}
where $\Delta G_i^{2d}\in \mathbb{R}^{N\times 2\times H\times W}$ represents the offset of each element in the grid matrix, where $N$ represents the number of offset points, $2$ means offset in the height and width directions, respectively.

Based on the offset grid points in the three-dimensional frustum space and the offsets in the two-dimensional image space, it can be concluded that the coordinates of the sampling point in the final view pixel coordinates are $G_i^{2d}+\Delta G^{2d}_i$.

After deformable learning in two-dimensional space, the actual sampling points are the fusion of each element of the grid matrix and the offset. At the same time, the network realizes the aggregation of the image features corresponding to the sampling points by learning the weight of each sampling point, which is expressed as:
\begin{equation}
    Y_{i,k}(p_0) = \sum_{p_n\in \mathbb{R}} w(p_n) \cdot F_{i,k}(p_n)
\end{equation}
\begin{equation}
    p_n=p_0+\Delta p_n = G_i^{2d}(w_0, h_0)+\Delta G_i^{2d}(w_0,h_0)
\end{equation}
Among them, $Y_{i,k}$ is the $i$th source feature output in the $k$th stage after aggregating the sampling point features, and $p_n$ is the coordinates of the sampling point with offsets. $G_i^{2d}(w_0,h_0)$ represents the coordinates of the grid matrix $G_i^{2d}$ at the position $(w_0,h_0)$, $w(p_n)$ is the weight of the offset sampling point $p_n$, $F_{i,k}(p_n)$ is the pixel feature of the source view corresponding to the offset sampling point. 
After learning the offset, the coordinates of $p_n$ are generally a decimal, not an integer, and therefore do not correspond to the pixel coordinates on the feature map. 
Then, bilinear interpolation is required here to obtain the coordinates of the offset, expressed as:
\begin{equation}
\begin{aligned}
    &F_{i,k}(m) = \sum_{n}G_i^{2d}(n,m)\cdot F_{i,k}(n)\\
    &=\sum_{n}g^{2d}_i(n_x,m_x)\cdot g^{2d}_i(n_y,m_y)\cdot F_{i,k}(n)\\
    &=\sum_{n}max(0, 1-\Delta_x)\cdot max(0, 1-\Delta_y)\cdot F_{i,k}(n)
\end{aligned}
\end{equation}
where $\Delta_x = |n_x-m_x|$ and $\Delta_y = |n_y-m_y|$. 
After aggregating the sampling point features of a single view source image, the next step is to aggregate the image features from all views into reference features, as shown in Fig.~\ref{4-PSS}. The specific operation is:
\begin{equation}
    \hat{F}_{1,k} = \sum_{i=1}^N w_i\cdot Y_{i,k}
\end{equation}
Among them, $w_i\in \mathbb{R}^{N\times H\times W}$ is the weight that aggregates the features of each view. 
$Y_{1,k}$ and $\hat{F}_{1,k}$ represent the reference features before and after the $k$th stage of consistency aggregation respectively.
 
\begin{figure*}[!htbp]
    \begin{center}
    \includegraphics[width=0.6\linewidth]{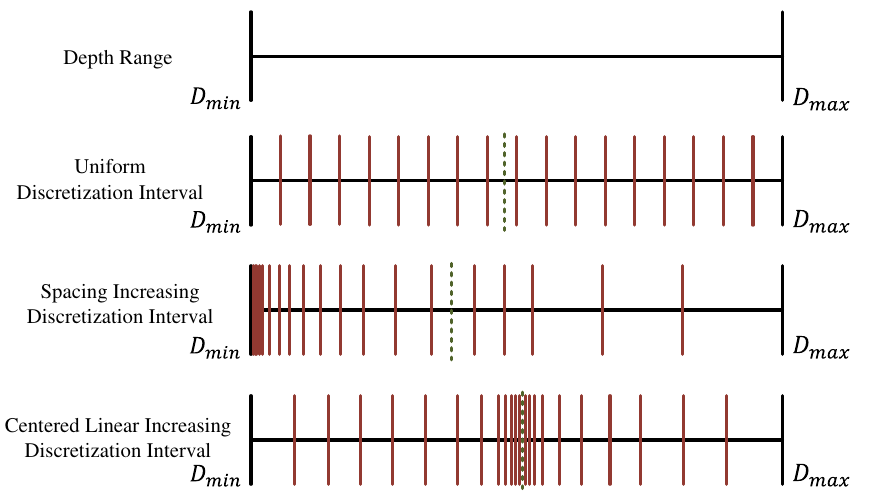}
    \end{center}
    \caption{Comparison of different discretization methods in the depth range. The first line in the figure represents the depth range, the second line denotes the Uniform Discretization (UD) interval, the third line denotes the Spacing Increasing Discretization (SID) interval, and the fourth line denotes the Centered Linear Increasing Discretization (CLID) interval proposed in this paper.}
    \label{5-DepthRange}
\end{figure*}

\subsection{Depth Hypothesis Deformable Discretization}
\label{ch4-sec-depth hypothesis deformable discretization}
The hypothesis of depth range and interval is a critical step in multi-stage multi-view stereo networks. 
The traditional depth hypothesis adopts uniform sampling and indents the depth range step by step. 
However, in complex actual scenes, the depth values corresponding to different pixels vary greatly, resulting in the common depth hypothesis being unable to adapt to various complex scenes, which is particularly prominent in large-scale remote sensing scenes. 
At the same time, if there are serious brightness unevenness problems in the actual data collection process, the traditional depth hypothesis will bring large errors to the depth estimation. 
To solve this problem, we adaptively construct the depth range and interval based on the predicted depth map of the previous stage through constructing a deformable discretization of the depth hypothesis and improving the rationality and accuracy of the depth hypothesis through performing deformable processing on the range and interval, respectively.  

The multi-scale feature extractor extracts deep features at three scales, corresponding to the three stages of subsequent depth estimation. In the $k$th stage, the depth features of the current stage are wrapped into multi-view image space based on the constructed depth hypothesis and the camera internal and external parameters and then a 3D cost volume is generated. 
The cost volume initially predicts the depth map $\mathcal{D}_k$ through a regression network and its corresponding depth probability map $P_k$. For the calculation of the depth map, we use the expected value of the depth distribution of each pixel to determine the depth estimate. Specifically, the depth hypothesis and depth probability value at pixel $x$ are $R_{k,i}(x)$ and $P_{k,i}(x)$, respectively. $d_k$ represents the number of depth hypothesis planes. $i$ indexes the depth hypothesis plane.
The estimated depth $D_k(x)$ of pixel $x$ can be calculated:
\begin{equation}
    D_k(x)=\sum_{i=1}^{d_k}R_{k,i}(x)\cdot P_{k,i}(x)
\end{equation}

In the $k+1$th stage, the estimated depth map of the $k$th stage is used as the central value of the depth hypothesis, and a new depth hypothesis is generated by constructing a depth hypothesis deformable strategy. First, based on the estimated depth value $\mathcal{D}_{k}$ at the $k$th stage, the depth hypothesis $R_{k,i}$ and the depth probability map $P_{k,i}$ are adopted to calculate the degree of dispersion of the depth hypothesis of the previous stage to the input depth value of the current stage, that is, calculate the variance of the depth hypothesis of the $k$th stage:
\begin{equation}
    S_k(x) = \sum_{i=1}^{d_k} P_{k,i}(x)\cdot (R_{k,i}(x) - \mathcal{D}_{k}(x))
\end{equation}
Among them, $S_k(x)$ is the variance value of the depth hypothesis at the $k$th stage pixel position $x$, and $i$ represents the $i$th layer depth hypothesis. Then the standard deviation value of the corresponding depth hypothesis is $\sigma_{k}(x)=\sqrt{S_k(x)}$.

Given the input depth map $\mathcal{D}_{k}(x)$ at the $k+1$th stage and the standard deviation of the depth hypothesis $\sigma_k(x)$, we first construct the upper limit and lower limit of the depth range which regards the predicted depth as the center of the range, aiming to achieve the deformable operation of the depth range. 
The distances of the upper and lower limits from the center are expressed as $+\eta \sigma_k(x)$ and $-\eta \sigma_k(x)$ respectively. On this basis, the entire depth range hypothesis $I_{k+1}(x)$ can be constructed:
\begin{equation}
    I_{k+1}(x) = [\mathcal{D}_k(x)-\eta \sigma_k(x), \mathcal{D}_k(x)+\eta \sigma_k(x)]
\end{equation}

On the basis of the deformable operation of depth range, to adapt the depth interval hypothesis of depth estimation in different environments, we introduce a depth interval deformation mechanism. Specifically, as shown in Fig.~\ref{5-DepthRange}, we divide linearly increasing discretization intervals toward the upper and lower limits of the interval by taking the estimated depth map as the center:
\begin{equation}
    R^{UP}_{k,d_i} = \mathcal{D}_k(x) + \frac{\eta \sigma_k(x)}{d_k/2(d_k/2+1)}\cdot d_i(d_i+1)
\end{equation}
\begin{equation}
    R^{DOWN}_{k,d_i} = \mathcal{D}_k(x) - \frac{\eta \sigma_k(x)}{d_k/2(d_k/2+1)}\cdot d_i(d_i+1)
\end{equation}
Among them, $R^{UP}_{k,d_i}$ and $R^{DOWN}_{k,d_i}$ are the depth values toward the upper limit and the lower limit respectively, and $d_k$ is the number of depth interval hypothesis. $d_i$ is the index of the depth interval plane.

\subsection{Loss function}
\label{ch4-sec-loss function}

In the multi-stage multi-view stereo task, the loss function consists of three parts, each part corresponding to the predicted output of the depth map of each stage. Following Cas-MVSNet\cite{gu2020cascade}, we use the summation of L1 loss between the stage-by-stage predicted depth map and the real depth map as the final loss function to drive the network for training:
\begin{equation}
     \mathcal{L}_{total} = \sum_{\lambda_i=1}^{3}\lambda_i \sum_{x\in\mathbf{x}_{valid}}||\mathcal{D}_{i }(x) - \hat{\mathcal{D}}(x)||_1
\end{equation}
Among them, $\lambda_i$ is the loss weight of each stage, $\mathcal{D}_i(x)$ and $\hat{\mathcal{D}}(x)$ represent the predicted depth value and ground truth of each stage, respectively.

\section{Experiments}
This section presents the results of experiments performed to demonstrate the effectiveness of our view space and depth deformable learning paradigm (SDL-MVS) in remote sensing multi-view stereo tasks.

\subsection{Datasets}\label{sec:Datasets} 
We use two large remote sensing multi-view datasets, LuoJia-MVS\cite{li2023hierarchical} dataset and WHU\cite{liu2020novel} dataset, for experimental verification.
 
The LuoJia-MVS dataset\cite{li2023hierarchical} is a large-scale aerial multi-view dataset. 
The dataset is DSM data in OSGB format constructed through a series of software such as Photoscan, Smart3D, and Meshmixer. Among them, the dataset is collected from part of Baiyun District, Guiyang City, Guizhou Province, China. 
This part of the area is a hilly basin area with mountains and hills as the main landforms, and its altitude ranges from approximately 1200$m$ to 1600$m$. 
The dataset covers a variety of land types, including cultivated land, forest land, urban areas, rural areas, industrial areas, mining areas, residential areas, unused land, etc. 
The processed data are combined into units in the five-view image set format. 
Each view data consists of an RGB image with a pixel size of $768\times 384$ and a resolution of $0.1m$ and its pixel-by-pixel depth map.

The WHU dataset\cite{liu2020novel} is also a large-scale aerial multi-view dataset, in which remote sensing multi-view images are captured by an oblique five-view camera device mounted on an unmanned aerial vehicle (UAV). 
One of the five-view cameras mounted on the drone is pointed vertically downward, and the optical axes of the other four cameras around it are tilted at 40°. 
Such a multi-view camera configuration can ensure that most of the information on the side facades of buildings can be captured in large-scale scenes. The data set is collected from Meitan County, Guizhou Province, China, within a geographical spatial range of approximately $6.7\times 2.2km^2$, and the ground resolution of the data is approximately $0.1m$. 
This part of the area has a large number of tall buildings, sparsely arranged factories, and densely arranged mountain forests. In addition, there are some bare land and river features in the area covered by the data.

\subsection{Experimental Setting}\label{sec:Experimental Setting}
 
\subsubsection{Implementation Details}\label{sec:Implementation Details}
The code of our work was developed based on PyTorch 1.7.1, and the training and testing phases were conducted on servers equipped with NVIDIA A40.
For all experimental processes, our work sets the batch size to 1 and uses the Adam optimizer\cite{kingma2017adam} with the parameters $\beta_1=0.9$ and $\beta_2=0.999$.
During the experiments, the number of training epochs was set to 24, and the initial learning rate was set to 0.001. 
During the training and testing process, the network uses three-view images and five-view images as input, respectively, to verify the robustness of the network and the impact of data from different perspectives.

As a multi-stage cascade network, our work follows the settings of the existing multi-stage network\cite{gu2020cascade}. The number of depth hypotheses in the three stages needs to be preset, which is $\{48,32,8\}$.
Consistent with existing multi-stage multi-view stereo networks, the resolutions of the multi-scale feature maps of the feature extractor are $\{1/16, 1/4, 1\}$ of the input image size, respectively. In the training stage, the depth estimation loss function weights in the three stages are set to $\{0.5, 1.0, 2.0\}$, respectively.

\subsubsection{Evaluation Metrics}\label{sec:Evaluation Metrics}
In multi-view stereo tasks, our work uses four metrics to jointly evaluate the quality of depth estimation of the network: mean absolute error (MAE), $<0.6$m, $<3$-interval, and completeness.
\textbf{Mean Absolute Error (MAE)} is a measure of the error of the estimated depth map. It is calculated by the average of the L1 distance between the estimated depth value and the true depth value and only calculates the distance within 100 depth intervals, to exclude extreme outliers.
\textbf{$<0.6$m} is a measure of the accuracy of the estimated depth map, which is obtained by calculating the percentage of pixels whose L1 error is less than the $0.6$m threshold.
\textbf{$<3$-interval} is an indicator to measure the accuracy of the estimated depth map, which is obtained by calculating the percentage of pixels whose L1 error is less than 3 depth intervals.
\textbf{Completeness} is an indicator describing completeness, obtained by calculating the percentage of pixels in the depth map with estimated depth values.

\subsection{Performance Analysis}\label{sec:Performance Analysis}
\subsubsection{Quantitative Analysis}We verified the superior performance of our SDL-MVS on the LuoJia-MVS\cite{li2023hierarchical} dataset and WHU\cite{liu2020novel} dataset, respectively.

\begin{table}[htb]
\caption{Performance analysis of depth estimation on the LuoJia-MVS dataset. `Three Views' and `Five Views' denote images from three views and five views as input to the network, respectively.
\textbf{Bold} represents the best performance.}\label{ch4-tab-LuoJia-MVSResults}
    \centering
    \begin{tabularx}{1.0\linewidth}{c|c|ccc}
    \toprule[1.5pt]
    Num. of Views & Model & MAE & $<$0.6m & $<$3-interval \\ 
    \midrule
    \multirow{9}{*}{Three views} & PatchmatchNet\cite{barnes2009patchmatch} & 0.252 & 92.7  & 87.2  \\
    ~ & Fast-MVSNet\cite{yu2020fast}   & 0.194 & 95.7 & 92.0  \\
    ~ & MVSNet\cite{yao2018mvsnet}        & 0.172 & 96.1 & 92.4  \\
    ~ & R-MVSNet\cite{yao2019recurrent}      & 0.177 & 96.0 & 93.5  \\
    ~ & RED-Net\cite{liu2020novel}       & 0.109 & 98.2 & 96.9  \\
    ~ & Cas-MVSNet\cite{gu2020cascade}    & 0.103 & 98.4 & 97.1  \\
    ~ & HDC-MVSNet\cite{li2023hierarchical}    & 0.089 & 98.7 & 97.8  \\
    ~ & \textbf{SDL-MVS (ours)}       & \bf{0.086}  & \bf{98.9} & \bf{98.9}   \\
    \midrule
    \multirow{9}{*}{Five views} & PatchmatchNet\cite{barnes2009patchmatch}  & 0.283 & 90.4 & 84.1 \\
    ~ & Fast-MVSNet\cite{yu2020fast}    & 0.357 & 84.6 & 74.9 \\
    ~ & MVSNet\cite{yao2018mvsnet}         & 0.270 & 91.2 & 81.8 \\
    ~ & R-MVSNet\cite{yao2019recurrent}       & 0.259 & 92.3 & 86.7 \\
    ~ & RED-Net\cite{liu2020novel}        & 0.156 & 94.9 & 90.5 \\
    ~ & Cas-MVSNet\cite{gu2020cascade}     & 0.141 & 97.9 & 95.4 \\
    ~ & HDC-MVSNet\cite{li2023hierarchical}     & 0.121 & 98.3 & 96.6 \\
    ~ & \textbf{SDL-MVS (ours)}        & \bf{0.118} & \bf{98.6} & \bf{98.2} \\ 
    \bottomrule[1.5pt]
    \end{tabularx}
\end{table}

\begin{table*}[htb]
\caption{Performance analysis of depth estimation on the WHU dataset. `Three Views' and `Five Views' represent images from three views and five views as input to the network, respectively.
\textbf{Bold} represents the best performance.}\label{ch4-tab-WHUDatasetResults}
    \centering
    \begin{tabularx}{0.6\linewidth}{c|c|cccc}
    \toprule[1.5pt]
    Number of Views & Model & MAE &$<0.6$m &$<3$-interval  & Comp \\ 
    \midrule
    \multirow{9}{*}{Three views} & PatchmatchNet\cite{barnes2009patchmatch} & 0.173 & 96.5 & 94.8 & 100\% \\
    ~ & Fast-MVSNet\cite{yu2020fast}   & 0.184 & 95.5 & 94.1 & 100\% \\
    ~ & MVSNet\cite{yao2018mvsnet}        & 0.190 & 95.0 & 94.3 & 100\% \\
    ~ & R-MVSNet\cite{yao2019recurrent}      & 0.183 & 95.3 & 93.5 & 100\% \\
    ~ & RED-Net\cite{liu2020novel}       & 0.112 & 98.1 & 97.9 & 100\% \\
    ~ & Cas-MVSNet\cite{gu2020cascade}    & 0.111 & 97.7 & 97.6 & 100\% \\
    ~ & HDC-MVSNet\cite{li2023hierarchical}    & 0.101 & 97.9 & 97.8 & 100\% \\
    ~ & \textbf{SDL-MVS (ours)}       & \bf{0.095} & \bf{98.4}  & \bf{98.0} & 100\% \\
    \midrule
    \multirow{9}{*}{Five views} & PatchmatchNet\cite{barnes2009patchmatch}  & 0.160 & 96.9 & 95.0 & 100\% \\
    ~ & Fast-MVSNet\cite{yu2020fast}    & 0.157 & 96.1 & 95.6 & 100\% \\
    ~ & MVSNet\cite{yao2018mvsnet}         & 0.160 & 95.8 & 95.5 & 100\% \\
    ~ & R-MVSNet\cite{yao2019recurrent}       & 0.173 & 95.4 & 93.8 & 100\% \\
    ~ & RED-Net\cite{liu2020novel}        & 0.104 & 98.1 & 97.9 & 100\% \\
    ~ & Cas-MVSNet\cite{gu2020cascade}     & 0.095 & 97.8 & 97.8 & 100\% \\
    ~ & HDC-MVSNet\cite{li2023hierarchical}     & 0.087 & 98.1 & 98.0 & 100\% \\
    ~ & \textbf{SDL-MVS (ours)}        & \bf{0.082} & \bf{98.7} & \bf{98.9} & 100\%  \\ 
    \bottomrule[1.5pt]
    \end{tabularx}
\end{table*}
\textbf{LuoJia-MVS dataset.}
The experiments compare the performance of our SDL-MVS with other methods on the LuoJia-MVS dataset. 
The comparison methods include PatchmatchNet\cite{barnes2009patchmatch}, Fast-MVSNet\cite{yu2020fast} , MVSNet\cite{yao2018mvsnet}, R-MVSNet\cite{yao2019recurrent}, RED-Net\cite{liu2020novel}, Cas-MVSNet\cite{gu2020cascade}, and HDC-MVSNet\cite{li2023hierarchical}.
Table~\ref{ch4-tab-LuoJia-MVSResults} gives the depth estimation error and accuracy performance of the proposed SDL-MVS paradigm on the LuoJia-MVS dataset.
Obviously, our SDL-MVS paradigm achieves the best depth estimation performance regardless of whether it takes three-view remote sensing images or five-view remote sensing images as input.
The mean absolute error (MAE) measures the difference between the estimated value and the true value and is an important indicator for evaluating estimation error.
In MAE, the proposed SDL-MVS paradigm achieves an estimation error of 0.086m in three-view estimation, which is \textbf{3\%} lower than the current state-of-the-art HDC-MVSNet\cite{li2023hierarchical} in terms of error (0.086 vs. 0.089). 
Compared with RED-Net\cite{liu2020novel}, the pioneering work of remote sensing multi-view stereo, the error is reduced by \textbf{21\%} (0.086 vs. 0.109). It can be seen that the proposed SDL-MVS paradigm achieves the best performance in estimation error and is better than all other methods. This also proves the effectiveness and superiority of deformable learning of views and depth in multi-view stereo.
In the accuracy evaluation, $<0.6$m and $<3$-interval are both indicators for measuring the proportion of accurately predicted pixels. The difference between the two is that they are measured from different angles. 
One is to set the actual distance threshold, and the other is to evaluate based on depth intervals.
It can be concluded from the experimental results that the proposed view space and depth deformable learning paradigm achieve the best performance in accuracy, reaching \textbf{98.9\%} of $<0.6$m and \textbf{98.9\%} of $<3$-interval respectively in three-view stereo task, which is a superior depth estimation performance.
Specifically, on $<3$-interval, the view space and depth deformable learning paradigm is \textbf{1.1\%} higher than the optimal HDC-MVSNet\cite{li2023hierarchical} method, which is a large improvement in multi-view depth estimation.
This once again proves the effectiveness of the deformable learning method in our work in improving depth estimation accuracy.
Since the LuoJia-MVS dataset tends to include scenes in both urban and natural areas, for large-scale complex scenes, the view space and depth deformable learning scheme of this work can better adapt to the depth estimation of multiple complex scenes.

\textbf{WHU dataset.}
The experiments compare the performance of our SDL-MVS paradigm with other methods on the WHU dataset, including PatchmatchNet\cite{barnes2009patchmatch}, Fast-MVSNet\cite{yu2020fast}, MVSNet\cite{yao2018mvsnet }, R-MVSNet\cite{yao2019recurrent}, RED-Net\cite{liu2020novel}, Cas-MVSNet\cite{gu2020cascade}, and HDC-MVSNet\cite{li2023hierarchical}.
The comparison of the depth estimation performance of our SDL-MVS paradigm and other methods is shown in Table \ref{ch4-tab-WHUDatasetResults}.
Through comparison, it was found that whether three-view remote sensing images or five-view remote sensing images were used as input, the view space and depth deformable learning paradigm proposed in our work achieved the best depth estimation performance of all other methods, reaching the lowest MAE error and highest accuracy.
Specifically, the proposed SDL-MVS scheme achieves an error of 0.082m in five
-view depth estimation, which is \textbf{5.7\%} lower than the state-of-the-art HDC-MVSNet\cite{li2023hierarchical} method (0.082 vs. 0.087), and in the five-view depth estimation task, the proposed method achieves \textbf{98.7\%} and \textbf{98.9\%} performance on the two accuracy indicators of $<0.6$m and $<3$-interval respectively, which is better than all other methods.

In addition, our SDL-MVS can achieve the best performance in three-view depth estimation. 
Because of the lack of the number of views when objects are occluded, the consistent information between views is not enough to support fine depth estimation, and the proposed progressive space deformable feature sampling method can realize the complementarity of information between multiple views, and effectively solves the occlusion problem between views when the number of views is insufficient.
Research can find that compared to the LuoJia-MVS dataset, the WHU dataset has many more buildings than the LuoJia-MVS dataset, which leads to the widespread occlusion problem in the WHU dataset. This aspect reflects the superiority of the proposed deformable learning scheme for solving occlusion problems.
The deformable discretization of the depth hypothesis can not only avoid the impact of uneven brightness but also adaptively compress the depth to achieve high-precision depth estimation with three views and five views as input.

\begin{figure*}[!htbp]
    \begin{center}
    \includegraphics[width=1.0\linewidth]{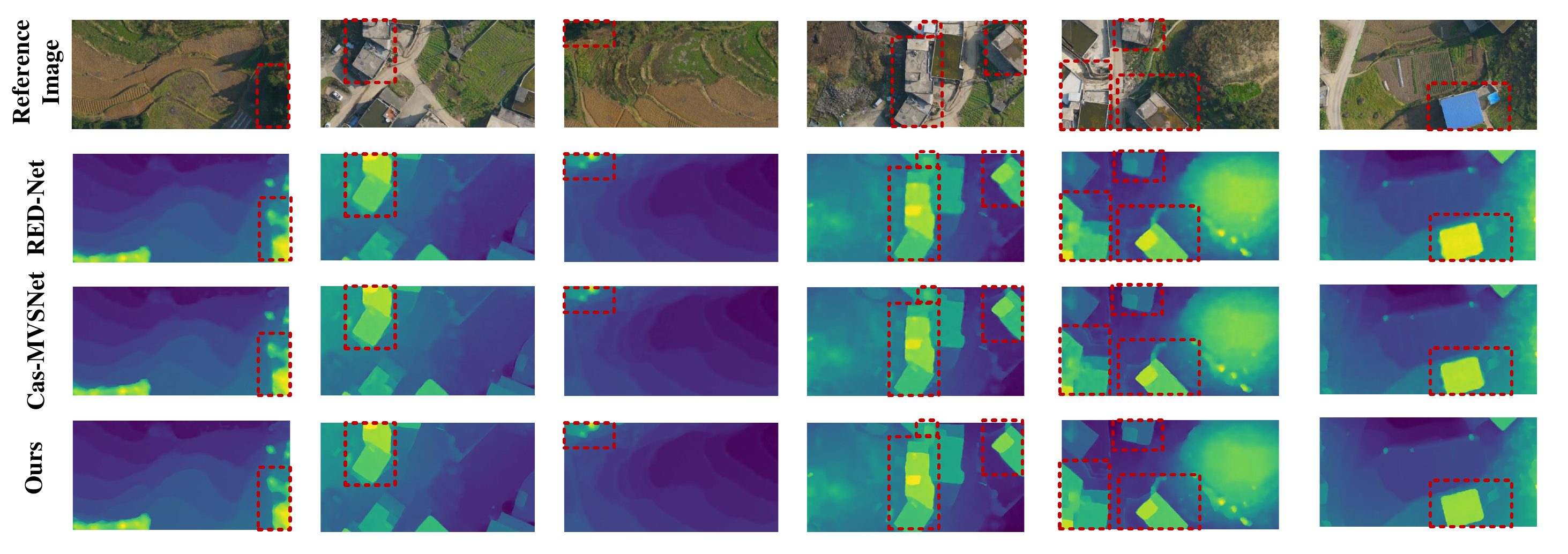}
    \end{center}
    \caption{Comparison of the depth estimation results of the proposed SDL-MVS and other methods on the LuoJia-MVS dataset. From top to bottom, results represent the input reference images, the depth estimation maps of RED-Net, the depth estimation maps of Cas-MVSNet, and the depth estimation maps of our SDL-MVS.}
    \label{6-LuoJiaDepthVisual}
\end{figure*}

\subsubsection{Qualitative Analysis}\label{sec:Qualitative Analysis}
To more intuitively demonstrate the superiority of the performance of our SDL-MVS, we provide the results of qualitative analysis on the LuoJia-MVS and WHU datasets and perform visualization analysis of the predicted depth maps, respectively.

\textbf{Visualizations of LuoJia-MVS dataset.}
To more intuitively demonstrate the prediction performance of the proposed SDL-MVS method, we provide the visualization results of the depth estimation of the proposed SDL-MVS on the LuoJia-MVS dataset.
As shown in Fig.~\ref{6-LuoJiaDepthVisual}, the first column in the figure is the center reference image input by the network, the second column is the depth estimation results of RED-Net\cite{liu2020novel}, the third column is the depth estimation results of Cas-MVSNet\cite{gu2020cascade}, and the fourth column is the result of depth estimation of the SDL-MVS method proposed in this work.
\begin{figure}[!htbp]
    \begin{center}
    \includegraphics[width=1.0\linewidth]{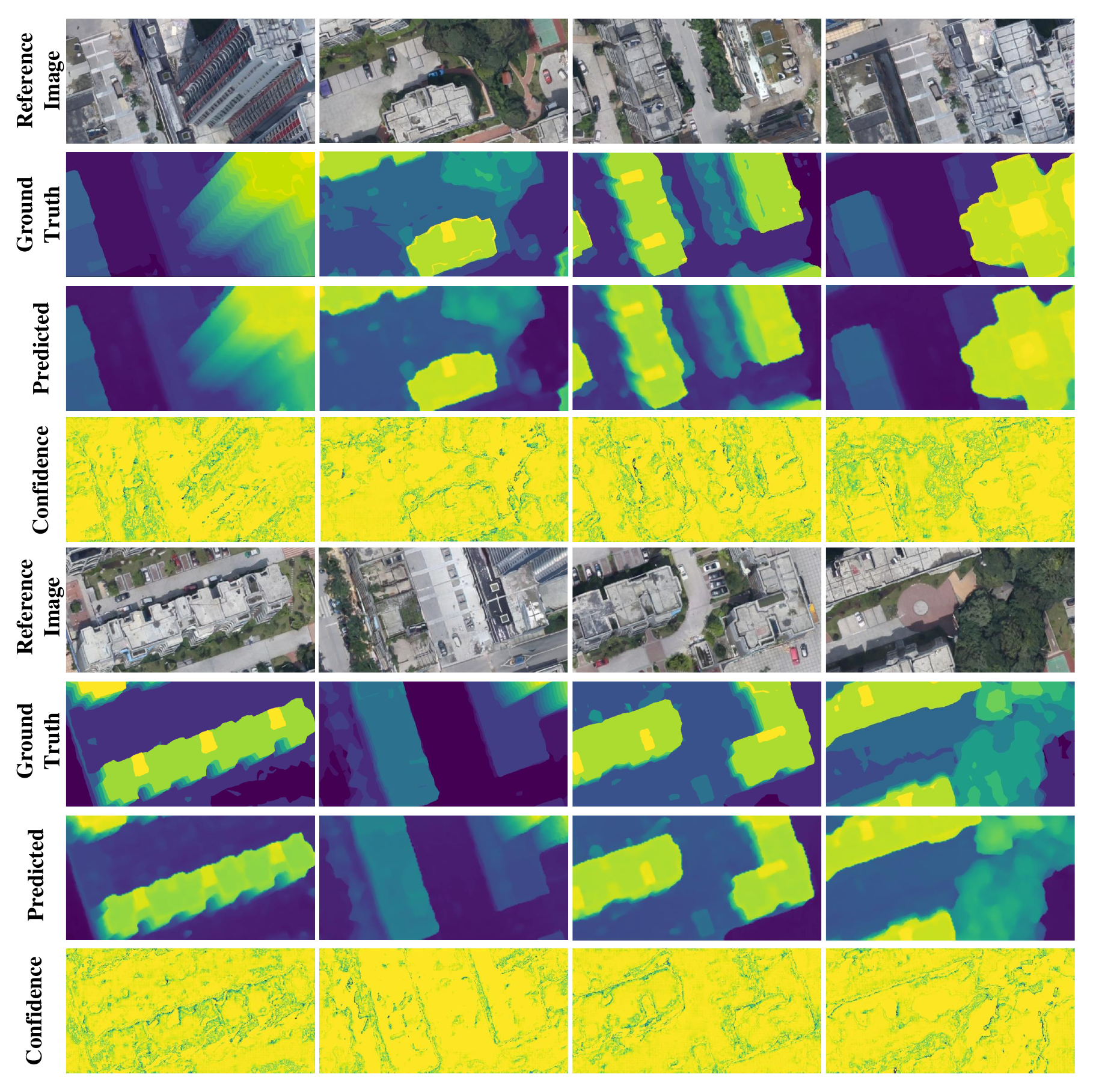}
    \end{center}
    \caption{Visualization analysis of the depth estimation results of the proposed SDL-MVS and the Ground Truth (GT) on the WHU dataset. The depth maps predicted by the proposed SDL-MVS are basically consistent with GT.}
    \label{7-WHUDepthVisual}
\end{figure}

To highlight the effectiveness of the proposed SDL-MVS in various scenarios, Fig.~\ref{6-LuoJiaDepthVisual} shows the results of depth estimation of the scenes containing only natural geographical instances and containing both natural geographical instances and buildings. 
In the depth estimation of physical geographical instances, the main instances are terraces and trees. However, terraces have a simpler structure than trees, and the irregular texture of trees causes certain difficulties in depth estimation for each method. However, compared with RED-Net and Cas-MVSNet, the depth estimation results in the figure can be seen that the depth predicted by the proposed SDL-MVS at the edge parts of trees and terraces is clear. 
The other two methods have the problem of prediction blur, which directly leads to the blur of the reconstructed point cloud.
This strongly proves that the proposed method of SDL-MVS has superior performance in prediction at edges because occlusions often appear at the boundaries of objects, which is consistent with the motivation of this work.

For building instances, the buildings in the red-marked area have relatively regular textures, so the depth predicted by the three methods has achieved superior results.
However, a horizontal observation shows that from RED-Net to Cas-MVSNet\cite{gu2020cascade} and then to the proposed SDL-MVS, the lines at the boundaries of the building are getting clearer and the prediction results are getting more and more refined.
In particular, for small houses with shorter roofs, there is certain obstruction from different views. As shown in the fourth row of the figure, RED-Net and Cas-MVSNet have blurry problems in the prediction of this occlusion scene. However, the deformable learning method proposed in this work can estimate its depth well and achieve superior results.

\textbf{Visualizations of WHU dataset.}
Similar to the LuoJia-MVS dataset, Fig.~\ref{7-WHUDepthVisual} shows the visualization results of the depth estimation of the proposed SDL-MVS on the WHU dataset.
Compared with the LuoJia-MVS dataset, the WHU dataset has the characteristics of dense building groups and sparse natural geographical instances such as trees, so the WHU dataset is more prone to occlusion problems.
In addition, the density of buildings makes the acquired multi-view image data extremely susceptible to light reflection.
Therefore, in the process of visualization, we specifically select areas with denser buildings for qualitative analysis. 
As shown in Fig.~\ref{7-WHUDepthVisual}, the prediction results of SDL-MVS proposed in this work on the WHU dataset are basically consistent with the ground truth, achieving superior depth estimation performance.
Especially in areas where trees and buildings meet, buildings are extremely susceptible to instances of irregular textures such as trees, but the proposed depth estimation network works well both in areas with dense buildings and in areas where buildings are mixed with irregular textures. 
All achieved impressive depth estimation performance and also showed excellent prediction results for depth estimation of high-rise buildings.

\begin{table}[!htbp]
\caption{Ablation experimental analysis of the proposed modules. `PSS' denotes Progressive Space deformable Sampling. `DHD Range' denotes Depth Range Hypothesis deformable Discretization. `DHD Interval' denotes Depth Interval Hypothesis deformable Discretization.}\label{ch4-tab-ablation_analysis}
    \centering
    \begin{tabularx}{1.0\linewidth}{ccc|ccc}
    \toprule[1.5pt]
    PSS & DHD Range & DHD Interval & MAE & $<3$-interval & $<0.6$m  \\ 
    \midrule
        -    &    -    &        -    & 0.135 & 96.4 & 96.8 \\
    $\surd$  &    -    &        -    & 0.118 & 97.2 & 97.7 \\
        -    & $\surd$ &        -    & 0.129 & 96.9 & 97.0 \\
        -    &     -   &    $\surd$  & 0.125 & 96.9 & 97.4 \\
    $\surd$  & $\surd$ &        -    & 0.096 & 97.5 & 98.0 \\
    $\surd$  &     -   &    $\surd$  & 0.100 & 97.9 & 98.3 \\
    $\surd$  & $\surd$ &    $\surd$  & \bf{0.095} & \bf{98.0} & \bf{98.4}  \\
    \bottomrule[1.5pt]
    \end{tabularx}
\end{table}

\subsection{Ablation Studies}\label{sec:Ablation Studies}
To verify the effectiveness of our view space and depth deformable learning paradigm, we conduct a series of ablation studies to investigate the impact of each component on multi-view stereo.

\subsubsection{Effectiveness of Progressive Space Deformable Sampling}
Progressive space deformable sampling can effectively reduce the impact of inevitable feature noise from different views, including noise from occluded features and pixel noise caused by uneven brightness, effectively improving the performance of depth estimation in multi-view stereo.
Table~\ref{ch4-tab-ablation_analysis} gives the effectiveness verification of the proposed progressive space deformable sampling mechanism.
On the basis of the baseline model, after using progressive space deformable sampling, the method proposed in this work has significant performance improvements in terms of error and accuracy, from 0.135m MAE error, 96.4\% $<3$-interval and 96.8\% $<0.6$m to 0.118m MAE, 97.2\% $<3$-interval and 97.7\% $<0.6$m.
The performance improvement is due to the superiority of the proposed progressive space deformable feature sampling, which can learn deformable sampling points in three-dimensional space and two-dimensional space between multi-view image features, and thus can adaptively adjust the coordinate positions of the sampling points before and after the three-dimensional to two-dimensional space projection. 
By adaptively searching for occluded object features or similar features around the occluded object, sampling is performed to reduce the impact of the noise of occluded features, bringing great improvement to the performance of the model.

\subsubsection{Effectiveness of Depth Range Hypothesis Deformable Discretization}
The deformability of the depth range hypothesis helps to set a more accurate depth range prior information at each stage, promotes the subsequent regression of accurate cost volume features and improves the overall depth estimation accuracy.
On the basis of the baseline model, after using the depth range hypothesis deformable discretization in the depth hypothesis, our method achieves more refined compression of the depth range hypothesis based on the predicted depth map of the previous stage, which greatly improves the depth estimation in the next stage. 
Better priors have improved the performance to the current MAE error of 0.129m, 96.9\% of $<3$-interval, and 97.0\% of $<0.6$m accuracy. The reason is that multi-stage multi-view stereo depth estimation achieves progressive optimization of depth from coarse to fine through the progression between stages, and the deformable range hypothesis can adaptively adjust the fine compression of the depth range.
Since the initially set depth hypothesis has a larger range, this is suitable for depth estimation of low-frequency features at small resolutions. As the stages increase, the reduction of the depth range helps to perform fine learning of high-frequency features at large resolutions, which reflects the importance of range uncertainty for overall depth estimation.
In particular, on the basis of progressive space deformable sampling, the addition of the deformable depth range hypothesis makes the depth estimation performance improve from the original MAE error of 0.118m, 97.2\% $<3$-interval and 97.7\% $<0.6$m accuracy has been improved to the current MAE error of 0.096m, 97.5\% $<3$-interval and 98.0\% $<0.6$m accuracy, which further proves the effectiveness of depth range hypothesis deformable strategy.

\subsubsection{Effectiveness of Depth Interval Hypothesis Deformable Discretization}
Depth interval hypothesis deformable discretization helps to adaptively adjust the sampling intervals above and below the estimated depth, which facilitates searching in the neighborhood around the depth prediction value, thereby improving the performance of stage-by-stage depth estimation.
The depth hypothesis not only needs to set the range of the depth space but also needs to set the interval of the depth hypothesis. 
The existing methods only consider the setting of the range and ignore the importance of the interval hypothesis for the entire depth estimation task.
Previous methods are based on the hypothesis of uniform intervals, but this setting does not meet the requirements of actual depth estimation. In remote sensing scenarios, due to the large acquisition range, uniform depth intervals on depth rays may cause the features on the source image to tend to be consistent, resulting in redundant depth intervals. After adding depth interval hypothesis deformable discretization, the performance of depth estimation improves to the current MAE error of 0.125m, 96.9\% of $<3$-interval, and 97.4\% of $<0.6$m accuracy.
Furthermore, after adding progressive space deformable sampling, introducing depth interval hypothesis deformable discretization makes the depth estimation performance improve from the original MAE error of 0.118m, 97.2\% $<3$-interval and 97.7\% $<0.6$m accuracy to the current MAE error of 0.100m, 97.9\% $<3$-interval, and 98.3\% $<0.6$m accuracy. 
This further proves that the depth interval hypothesis has a certain influence on coarse-to-fine depth optimization.

In general, as shown in Table~\ref{ch4-tab-ablation_analysis}, after adding progressive space deformable sampling, depth range and interval hypothesis deformable discretization, the overall depth estimation performance is improved from the original 0.135m MAE error, 96.4\% $<3$-interval and 96.8\% $<0.6$m accuracy to the current MAE error of 0.095m, 98.0\% $<3$-interval and 98.4\% $ <0.6$m accuracy. 
This verifies the effectiveness and superiority of the proposed progressive space deformable sampling and depth hypothesis deformable discretization.
\begin{table}[!htbp]
\caption{Experimental analysis of deformable point generation methods. `Random Point' denotes random initialization of deformable points. `Convolution Kernel Point' denotes a deformable point initialized through the convolution kernel.}\label{ch4-tab-samplingpoint}
    \centering
    \begin{tabular}{c|ccc}
    \toprule[1.5pt]
    Deformable Point Generation & MAE &$<3$-interval &$<0.6$m \\ 
    \midrule
    Random Point             & 0.095 & 98.0 & 98.4 \\
    Convolution Kernel Point & 0.101 & 97.8 & 98.3  \\
    \bottomrule[1.5pt]
    \end{tabular}
\end{table}

\subsection{Deep Analysis}
\subsubsection{Deformable Point Generation}
In three-dimensional frustum and two-dimensional image spaces, the way deformable points are generated greatly affects the learning of point offsets. 
Since the defined grid matrix is regularly arranged according to length and width (the depth dimension is added to the three-dimensional frustum space), there are two options when performing discrete sampling point learning. One is to use random points to perform within the neighborhood. The second way is to generate sampling points by using the points corresponding to the convolution kernel to offset them as the generated sampling points. The biggest difference between the two is the initialization position of the point. 
To determine the final sampling point learning method, we conduct experiments on random point generation and convolution kernel sampling point generation respectively. 
Table~\ref{ch4-tab-samplingpoint} gives the analysis results of the experiment. 
From the data in the table, it can be concluded that both the random point generation scheme and the convolution kernel sampling point generation scheme have superior performance, and the performance difference is small, but the random point generation shows a slight advantage. Therefore, in the actual experimental process, we adopt a random point generation scheme to learn sampling points.

\begin{table}[!htbp]
\caption{Experimental analysis of progressive space deformable sampling. `3D space' and `2D space' represent deformable feature sampling in 3D and 2D space respectively.}\label{ch4-tab-progressivespatialdeformable}
    \centering
    \begin{tabular}{cc|ccc}
    \toprule[1.5pt]
    3D space & 2D space & MAE &$<3$-interval &$<0.6$m \\ 
    \midrule
    $\surd$  &    -    & 0.097 & 97.8 & 98.0 \\
       -     & $\surd$ & 0.116 & 97.3 & 97.6 \\
    $\surd$  & $\surd$ & 0.095 & 98.0 & 98.4 \\
    \bottomrule[1.5pt]
    \end{tabular}
\end{table}

\subsubsection{Sampling Space}
In actual multi-view stereo tasks, the spaces involved include three-dimensional frustum space and two-dimensional image space. The grid matrix of the three-dimensional frustum space and the grid matrix of the two-dimensional space are generated before and after projection using the camera pose matrix between the views, respectively. 
To verify the effectiveness of sampling point generation in three-dimensional frustum space and two-dimensional space, we conduct experiments on progressive space sampling. As shown in Table~\ref{ch4-tab-progressivespatialdeformable}, we respectively generate deformable points in only three-dimensional frustum space, only two-dimensional space, and both three-dimensional and two-dimensional spaces (that is, the generation of progressive space deformable points).
Experimental results show that although using only a single space to generate deformable points can improve the depth estimation performance, deformable points that exist in two spaces at the same time have more advantages in performance. 
In addition, the experimental results strongly prove that the introduction of deformable points in three-dimensional frustum space has performance advantages compared to two-dimensional space. The reason is that three-dimensional space can cover all view features, and the learning of deformable sampling points in this space can better adapt to the feature offset of different views. 
This is a sharp contrast from the sampling point learning in two-dimensional space which is limited to the current view. 
However, sampling in three-dimensional and two-dimensional spaces at the same time can achieve the best performance. 
This is because the two are interdependent and complementary in sampling point learning. 
Learning three-dimensional and two-dimensional space sampling points separately can achieve multi-dimensional sampling. 
Adaptive feature aggregation of view features improves the performance of depth estimation.
\begin{figure*}[!htbp]
    \begin{center}
    \includegraphics[width=1.0\linewidth]{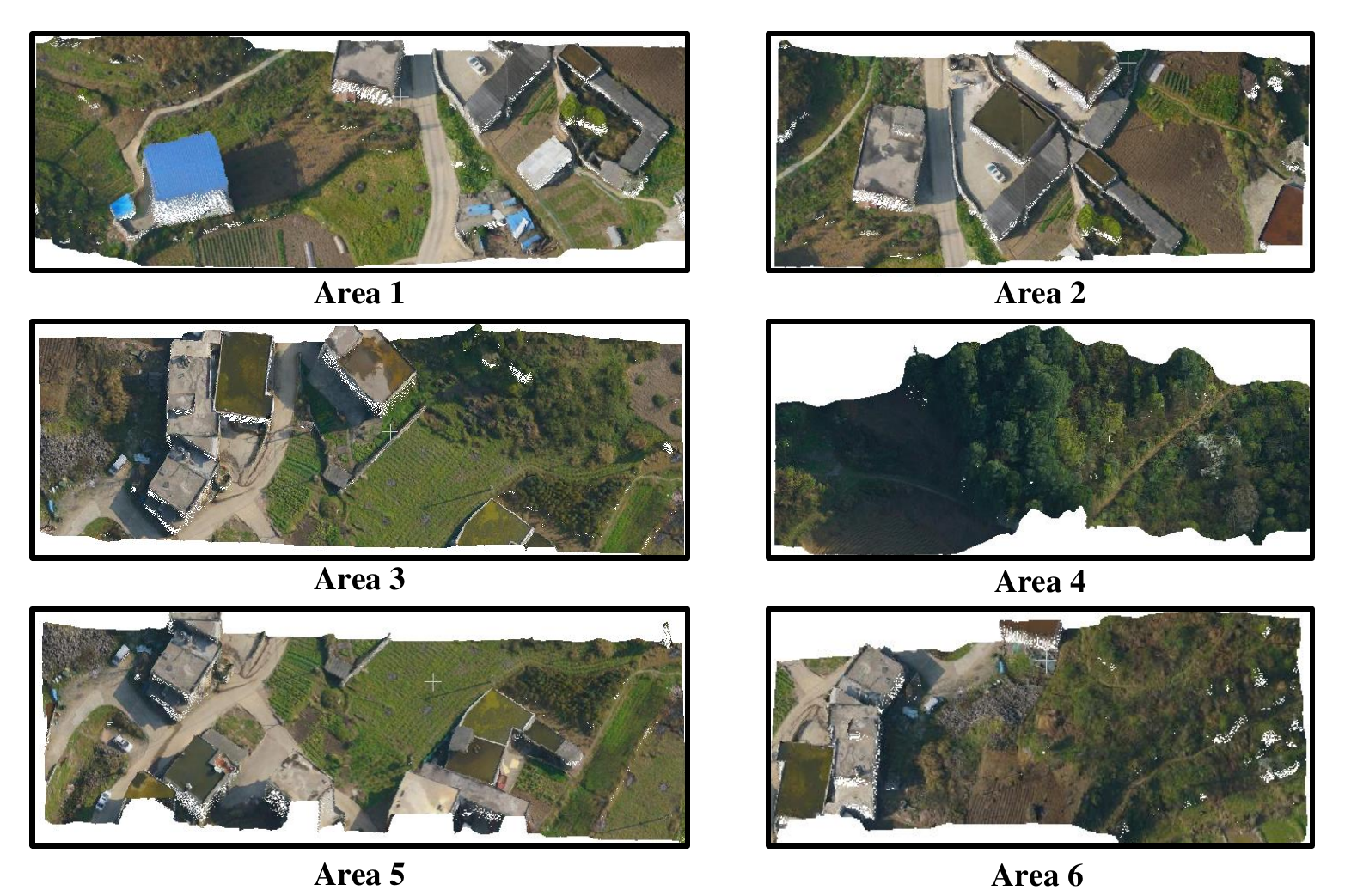}
    \end{center}
    \caption{Local visualization results of point cloud fusion of the proposed SDL-MVS on the LuoJia-MVS dataset. We selected 6 areas with large differences in instances for visual analysis.}
    \label{8-PointCloudVisualLuoJia}
\end{figure*}

\begin{table}[!htbp]
\caption{Experimental analysis of depth interval discretization methods. `UD' denotes Uniform Discretization. `SID' denotes Spacing Increasing Discretization. `CLID' denotes the proposed Centered Linear Increasing Discretization.}\label{ch4-tab-depth_interval}
    \centering
    \setlength\tabcolsep{11pt}
    \begin{tabularx}{1.0\linewidth}{c|ccc}
    \toprule[1.5pt]
    Depth Discretisation& MAE &$<3$-interval & $<0.6$m \\ 
    \midrule
    UD   & 0.096 & 97.5 & 98.0 \\
    SID  & 0.096 & 97.9 & 98.2 \\
    CLID & 0.095 & 98.0 & 98.4 \\
    \bottomrule[1.5pt]
    \end{tabularx}
\end{table}
\subsubsection{Deep Interval Discretisation Method}
In the depth interval discretization method, as shown in Fig.~\ref{5-DepthRange}, the solutions that can be adopted include uniform discretization (UD), spacing increasing discretization (SID), and centered linear increasing discretization (CLID). Existing methods focus on uniform discretization schemes to divide depth intervals, but this does not match the actual scenario. We conduct experimental verifications on three discretization schemes, as shown in Table~\ref{ch4-tab-depth_interval}. The experiment gives the depth estimation performance of three discretization schemes. Obviously, centered linear incremental discretization achieves the best depth estimation performance. However, according to the experimental results, it can be found that the performance produced by spacing increasing discretization and linear increasing discretization is not much different, because it is also a non-uniform discretization method. In summary, we finally selected the centered linear increasing discretization method as the interval discretization scheme of the network. 

\subsection{Visualization Analysis}
\subsubsection{Visualization Analysis of Point Cloud Fusion}
Point cloud data reconstructed from depth maps is specifically a process of using the predicted depth of pixels to map from the image coordinate system to the world coordinate system according to the internal and external parameter matrix of the camera. We fuse the depth maps of the test areas of the LuoJia-MVS dataset and the WHU dataset respectively, and perform visual analysis on the fused point cloud.

\begin{figure*}[!htbp]
    \begin{center}
    \includegraphics[width=1.0\linewidth]{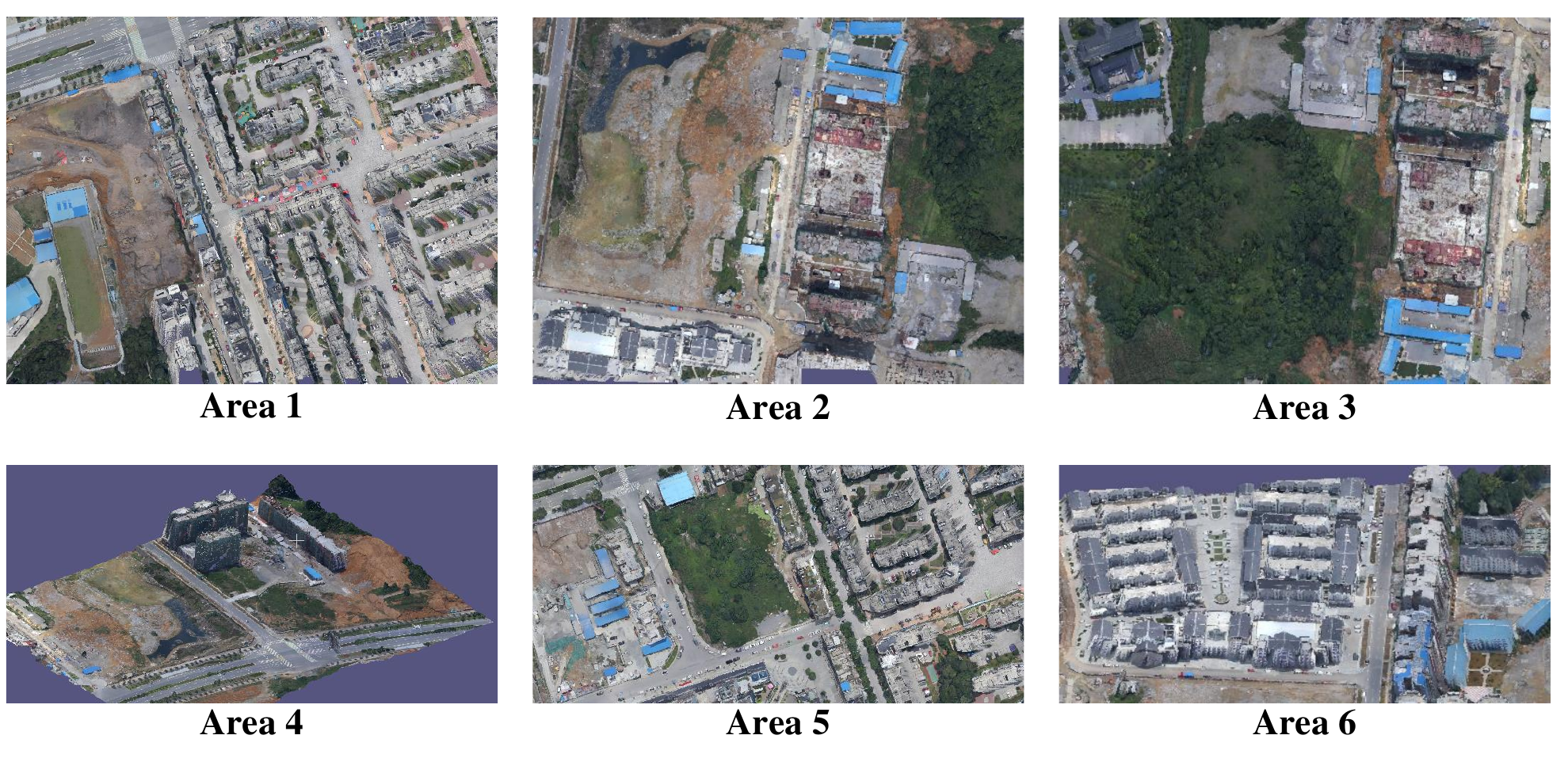}
    \end{center}
    \caption{Local visualization results of point cloud fusion of the proposed SDL-MVS on the WHU dataset. 
    The point cloud results are obtained by the fusion of multi-view depth maps. We selected areas with dense buildings and areas where buildings and vegetation coexist for visualization.}
    \label{9-PoinuCloudVisualPartWHU}
\end{figure*}

\begin{figure*}[!htbp]
    \begin{center}
    \includegraphics[width=1.0\linewidth]{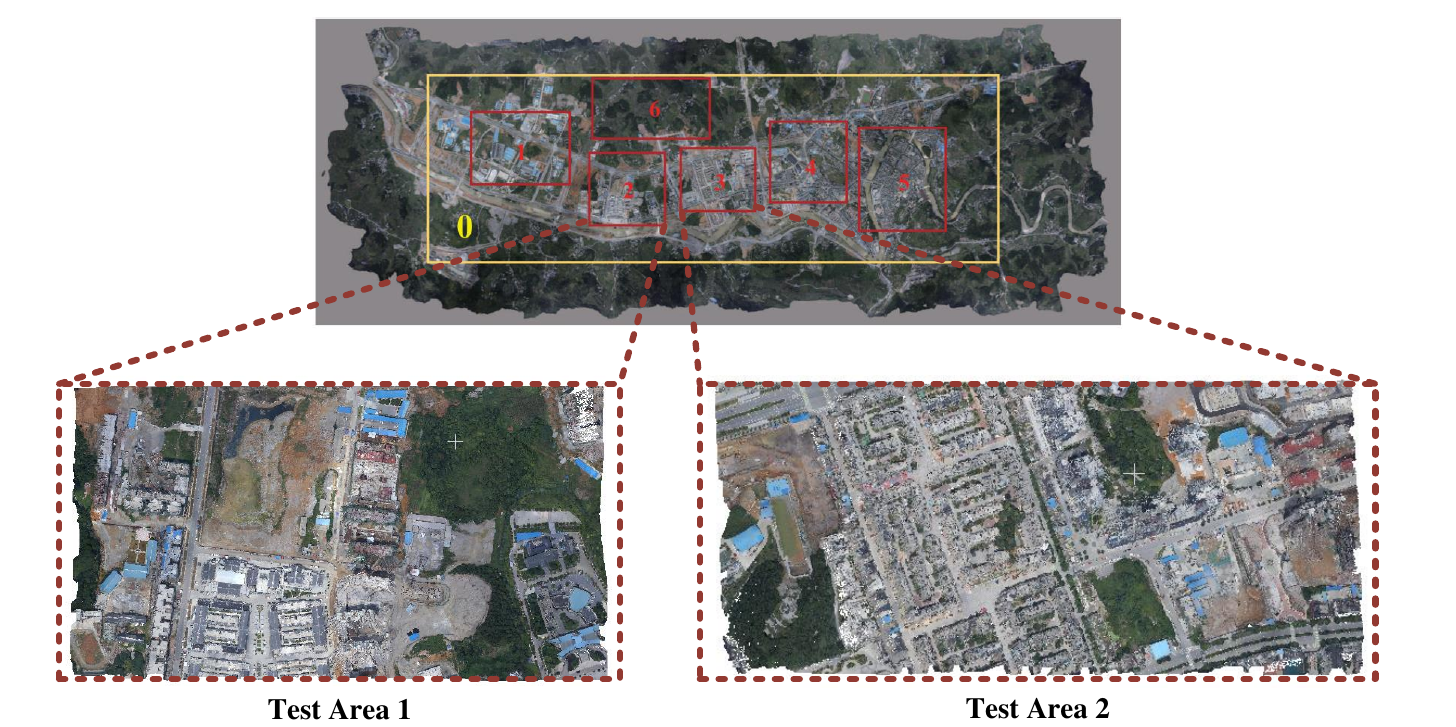}
    \end{center}
    \caption{Large-scale point cloud visualization results of the proposed SDL-MVS method on the WHU dataset. Point cloud fusion achieves superior results in large-scale scenes.}
    \label{10-PointCloudVisualLargeWHU}
\end{figure*}
\textbf{Point Cloud Fusion of LuoJia-MVS Dataset.}
The LuoJia-MVS dataset has the characteristics of both building groups and natural feature groups. 
To verify the point cloud reconstruction results from the perspective of qualitative analysis, we spatially map, fuse, and convert the depth map predicted by the proposed SDL-MVS into point cloud data in the world coordinate system.
As shown in Fig.~\ref{8-PointCloudVisualLuoJia}, we visualize the point cloud reconstruction results of 6 typical areas of the test set of the LuoJia-MVS dataset. 
Among them, Area 1, Area 3, Area 5, and Area 6 have large areas of buildings and natural geographical instances (trees, cultivated land, etc.). Area 2 has more buildings. Area 4 has a larger area of physical geographical instances (trees).
Buildings and natural geographical instances have large texture differences. 
Buildings are typical examples with regular textures, while natural geographical instances such as trees are typical examples with irregular textures.
In the field of multi-view reconstruction, instances of irregular textures similar to trees have always been the focus of researchers, and it has always been difficult to achieve high reconstruction accuracy.
However, from the visualization results shown in the figure, it can be concluded that the instances of irregular textures such as trees reconstructed in our work have impressive reconstruction effects, which also proves that SDL-MVS proposed in our work is effective and superior in the field of multi-view depth estimation.
In addition, for areas with regular textures such as buildings, the point cloud reconstructed in our work achieved fine reconstruction results, and partially occluded building areas also achieved good reconstruction results, which once again proved the effectiveness of deformable learning for depth estimation.
The point cloud reconstruction results of areas with different characteristics demonstrate the high accuracy and superiority of the depth estimation of the proposed view space and depth deformable learning.

\textbf{Point Cloud Fusion of WHU Dataset.}
Different from the LuoJia-MVS dataset, the collection areas of the WHU dataset mostly focus on areas where urban building groups are located, so there is less data collection of natural geographical instances such as trees.
However, to more convincingly prove the superiority of SDL-MVS proposed in our work in depth estimation, we select six regions of different scenes for visual analysis of point cloud reconstruction.
As shown in Fig.~\ref{9-PoinuCloudVisualPartWHU}, we visualize the point cloud reconstruction results of 6 typical areas of the test set of the WHU dataset. Among them, Area 3 and Area 4 have a large area of natural geographical instances, Area 2 has both a large area of building groups and natural geographical instance groups, and Area 1, Area 5 and Area 6 have gathered large-scale building groups.
Through qualitative analysis, it can be seen that for dense areas of building instances with regular textures, the view space and depth deformable multi-view stereo reconstruction method proposed in our work has achieved superior point cloud fusion and reconstruction results. In addition, especially for examples such as trees with irregular textures in large areas, it is difficult to reconstruct them with a small number of multi-view images. However, the multi-view stereo reconstruction method proposed in our work has achieved good results in areas with dense trees. The performance strongly proves the superiority of the proposed depth estimation method.

As shown in Fig. \ref{10-PointCloudVisualLargeWHU}, the upper part is the overall area of data collection, and the lower part is the large-scale point cloud reconstruction result of the test set. The reconstructed area 1 and area 2 correspond to Area 2 and 3 of the test set, respectively. Through qualitative analysis, it can be found that in large-scale reconstruction, refined 3D point cloud reconstruction of city-level areas can be achieved based on a small number of multi-view images, and the reconstruction results have achieved impressive results.

\section{Conclusion}

To solve the problem of blurred details caused by occlusion and uneven brightness between views, we proposed a multi-view stereo reconstruction method rooted in a view space and depth deformable learning (SDL-MVS) paradigm, which aims to learn the deformable interactions of features in different spaces for high-precision depth estimation.
To solve the problem of occlusion and uneven brightness, we proposed a progressive space deformable sampling (PSS) mechanism, which is based on the multi-view three-dimensional frustum space and two-dimensional image space to carry out the learning process of deformable sampling points in a progressive manner.
To further optimize the depth, we introduce a depth hypothesis deformable discretization (DHD) mechanism which can be adaptive to construct the depth range and interval hypothesis.
Finally, the network achieves explicit modeling of the two major problems of occlusion and uneven brightness faced in multi-view stereo through the view space and depth deformable learning paradigm, achieving more accurate multi-view depth estimation.
The experimental results on the LuoJia-MVS dataset and WHU dataset prove the effectiveness and superiority of the proposed SDL-MVS and achieve the best performance of depth estimation.

\bibliographystyle{IEEEtran}
\bibliography{egbib}
 
\end{document}